\documentclass[10pt,twocolumn,letterpaper]{article}

\usepackage{cvpr}              % To produce the CAMERA-READY version
\usepackage{times}
\usepackage{bm}
\usepackage{microtype}
\usepackage{epsfig}
\usepackage{caption}
\usepackage{float}
\usepackage{placeins}
\usepackage{color, colortbl}
\usepackage{stfloats}
\usepackage{enumitem}
\usepackage{tabularx}
\usepackage{xstring}
\usepackage{multirow}
\usepackage{xspace}
\usepackage{url}
\usepackage[hang,flushmargin]{footmisc}
\usepackage{algorithmic}
\usepackage{algorithm}
\usepackage{footnote}
\usepackage[dvipsnames]{xcolor}

\usepackage[accsupp]{axessibility}
\definecolor{cvprblue}{rgb}{0.21,0.49,0.74}
\usepackage[pagebackref,breaklinks,colorlinks,citecolor=cvprblue]{hyperref}
\usepackage{amsthm,amsmath,amssymb}
\usepackage{hyperref}
\usepackage{mathrsfs}

\def\paperID{1665} % *** Enter the Paper ID here
\def\confName{CVPR}
\def\confYear{2025}

\def\eg{\emph{e.g., }}
\title{\ourmethod: High Fidelity Scene Text Synthesis}
\newcommand{\ourmethod}{{\textsc {DreamText}}\xspace}
\author{
	\fontsize{12}{14}\selectfont
	\textbf{Yibin Wang}\textsuperscript{1,2},
	\textbf{Weizhong Zhang}\textsuperscript{1,4},
    \textbf{Honghui Xu}\textsuperscript{5},
	\textbf{Cheng Jin}\textsuperscript{1,3}\footnotemark[2]\\
	\textsuperscript{1}Fudan University 
        \textsuperscript{2}Shanghai Innovation Institute\\
        \textsuperscript{3}Innovation Center of Calligraphy and Painting Creation Technology, MCT \\
        \textsuperscript{4}Shanghai Key Laboratory of Intelligent Information Processing
        \textsuperscript{5}Zhejiang University of Technology \\
	{\tt\small yibinwang1121@163.com, weizhongzhang@fudan.edu.cn, xhh@zjut.edu.cn, jc@fudan.edu.cn} \\
}

\begin{document}

\twocolumn[{%
	\renewcommand\twocolumn[1][]{#1}%
	\maketitle
	\begin{center}
		\centering
		\captionsetup{type=figure}
		\includegraphics[width=0.95\linewidth]{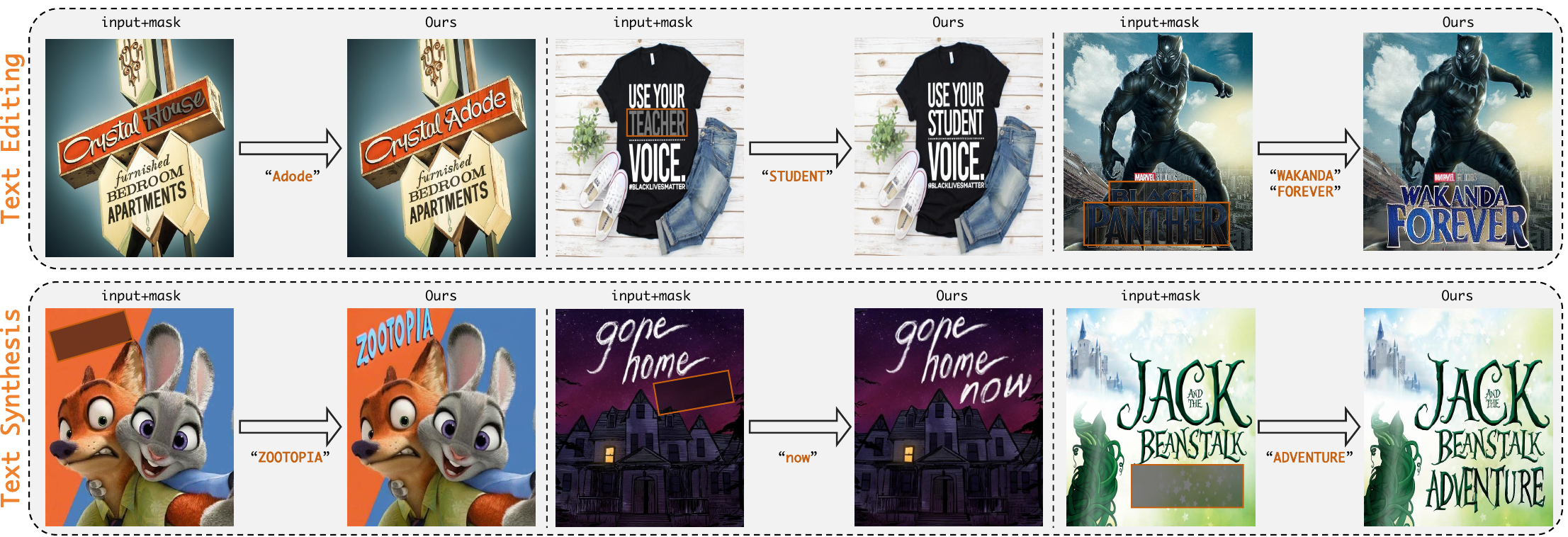}
		\captionof{figure}{Displayed are the results generated using our \ourmethod, showcasing its prowess across varied inputs.}
		\label{fig:teaser}
	\end{center}
}]
\renewcommand{\thefootnote}{}
\footnotetext{$^{\dag}$Corresponding author.}
\renewcommand{\thefootnote}{\arabic{footnote}}

\begin{abstract}
Scene text synthesis involves rendering specified texts onto arbitrary images. 
Current methods typically formulate this task in an end-to-end manner but lack effective character-level guidance during training.
Besides, their text encoders, pre-trained on a single font type, struggle to adapt to the diverse font styles encountered in practical applications.
Consequently, these methods suffer from character distortion, repetition, and absence, particularly in polystylistic scenarios.
To this end, this paper proposes \ourmethod for high-fidelity scene text synthesis.
Our key idea is to reconstruct the diffusion training process, introducing more refined guidance tailored to this task, to expose and rectify the model's attention at the character level and strengthen its learning of text regions.
This transformation poses a hybrid optimization challenge, involving both discrete and continuous variables. To effectively tackle this challenge, we employ a heuristic alternate optimization strategy. Meanwhile, we jointly train the text encoder and generator to comprehensively learn and utilize the diverse font present in the training dataset. This joint training is seamlessly integrated into the alternate optimization process, fostering a synergistic relationship between learning character embedding and re-estimating character attention.
Specifically, in each step, we first encode potential character-generated position information from cross-attention maps into latent character masks. These masks are then utilized to update the representation of specific characters in the current step, which, in turn, enables the generator to correct the character's attention in the subsequent steps.
Both qualitative and quantitative results demonstrate the superiority of our method to the state of the art. 
Our project page is \href{https://codegoat24.github.io/DreamText/}{here}.
\end{abstract}

\begin{figure*}[htb]
\centering
\includegraphics[width=0.9\textwidth]{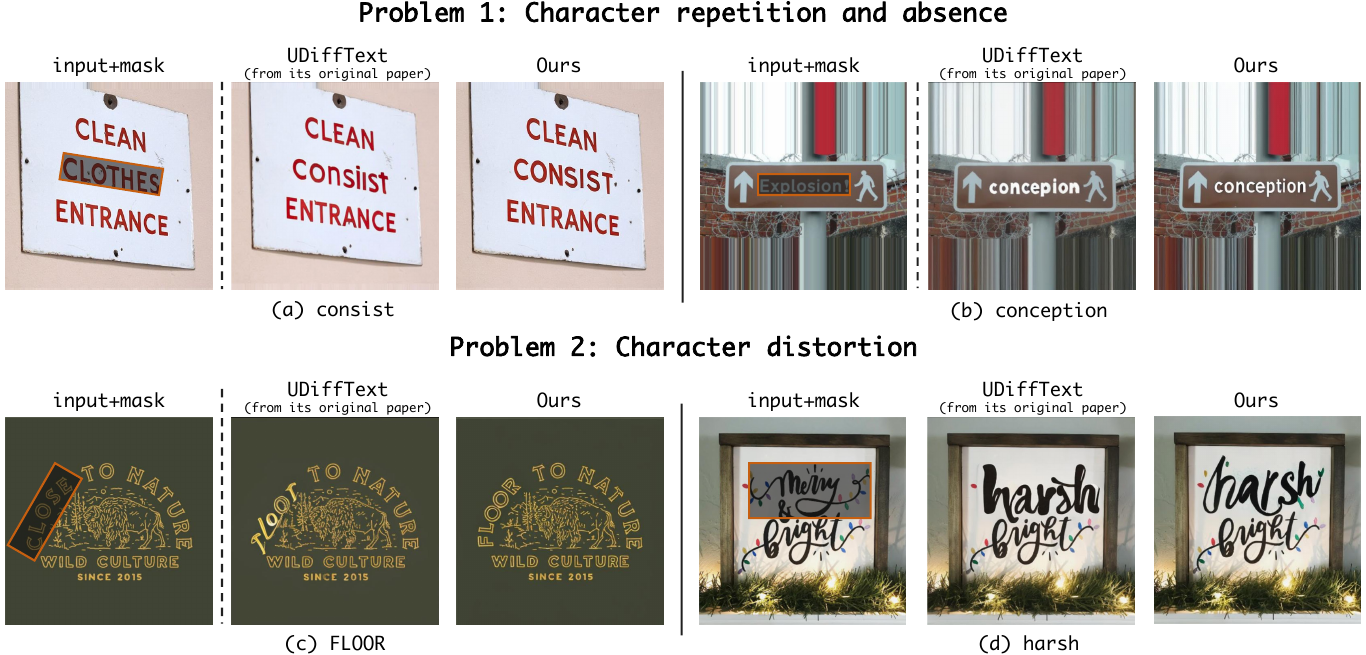}
\caption{Current methods encounter significant challenges, i.e., character repetition and absence (top) and character distortion (bottom).}
\label{fig:problem}
\vspace{-0.4cm}
\end{figure*}
\section{Introduction}
Scene text synthesis involves modifying or inserting specified text in arbitrary images while maintaining its natural and realistic appearance. Earlier GAN-based style transfer methods \cite{MOSTEL,park2021multiple,roy2020stefann,yang2020swaptext} accomplish this by transferring the text style from a reference image to the rendered target text image. 
However, they are constrained in their capacity to generate text in arbitrary styles, such as font and color. 
These limitations are effectively addressed by utilizing the diffusion model \cite{saharia2022photorealistic,DiffSTE,TextDiffuser,lakhanpal2024refining,liu2024glyph,wang2024magicface,wang2023enhancing}, harnessing its inherent prior knowledge and robust capabilities.
Nevertheless, these methods still encounter challenges in generating accurate text within polystylistic images due to the inadequate conditional guidance provided by their character-unaware text encoder. 
To this end, a recent study \cite{UdiffText} proposes a character-level text encoder, providing more robust conditional guidance for the diffusion model.

\begin{figure}[ht]
\centering
\includegraphics[width=1\linewidth]{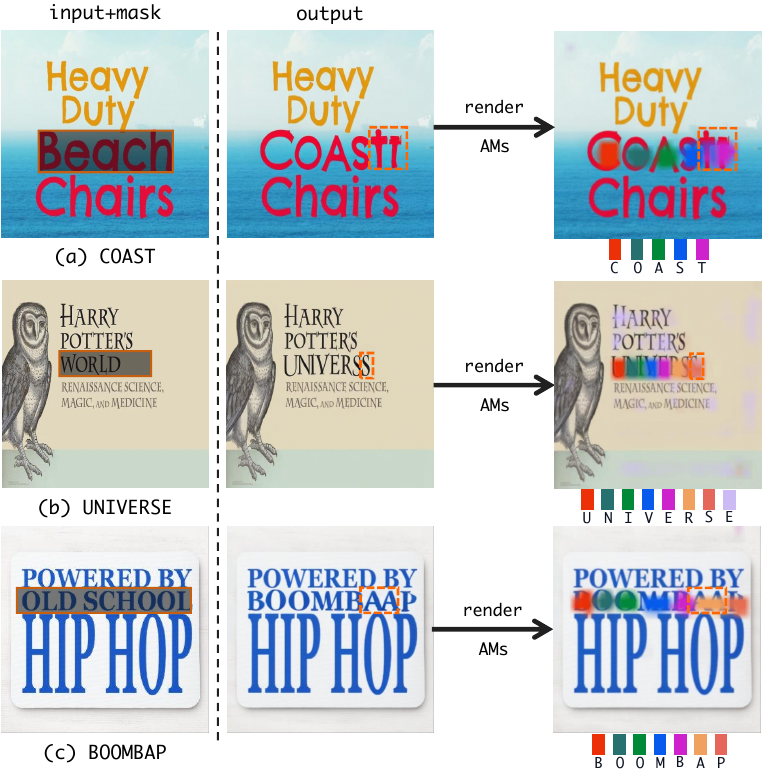}
\caption{The problematic results rendered by characters' attention maps (AMs).}
\label{fig:visual_problem}
\vspace{-0.4cm}
\end{figure}

\begin{figure}[t]
\centering
\includegraphics[width=1\linewidth]{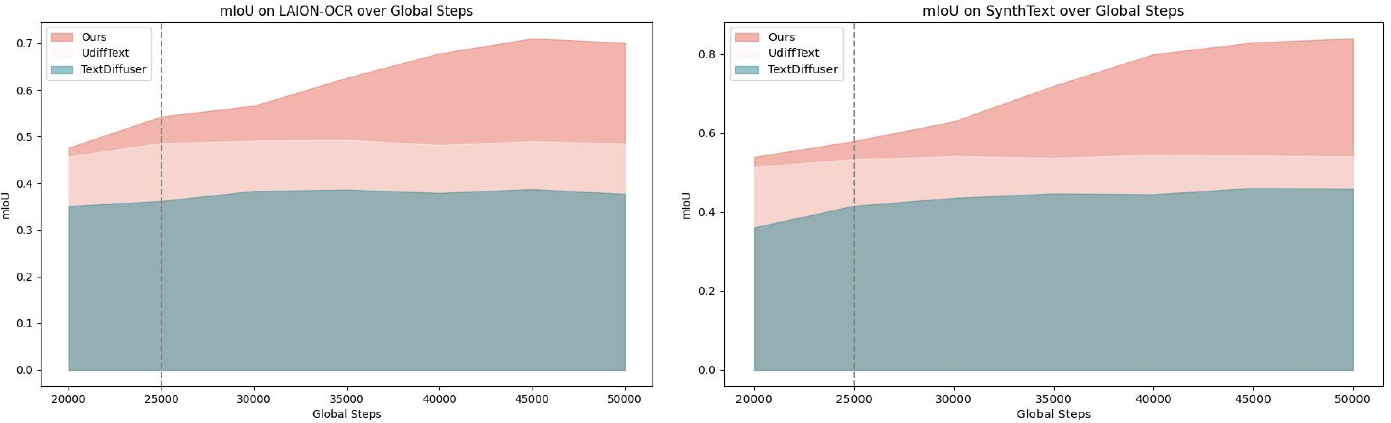}
\caption{The mIoU scores of UDiffText, TextDiffuser, and our method on LAION-OCR and SynthText over global training steps. Our method adopts a balanced supervision strategy: we initially use latent character masks to steer the character's attention for a warm-up in the earlier stage and stop guiding after 25,000 steps.}
\label{fig:mIoU}
\vspace{-0.5cm}
\end{figure}

Despite their effectiveness, these methods are still difficult to accurately render text within complex scenes.
Specifically, these methods such as UDiffText \cite{UdiffText} and TextDiffuser \cite{TextDiffuser} solely utilize a single font type to pre-train the text encoder, which is subsequently employed to fine-tune the generator.
However, the diverse font styles encountered in practical applications present a significant challenge: the restricted representation domain hampers their ability to render text with unseen font styles, \eg cases shown in Fig. \ref{fig:problem} (bottom). 
Furthermore, these methods rely on character segmentation masks to supervise characters' attention for position control. We argue this approach has significant limitations: (1) unlike classification tasks with uniquely determined labels, the optimal generation positions for characters can be diverse. Therefore, using the specific masks to rigidly constrain the model may limit its flexibility in estimating optimal positions, hindering its ability to adapt to varied and complex scenarios. Besides, (2) these masks generated by a pre-trained segmentation model \cite{TextDiffuser} are imprecise, tending to over-segment character regions.
As a result, the model encounters issues of character repetition and absence as illustrated in Fig. \ref{fig:problem} (top left) and (top right) respectively. 
We suspect these issues stem from characters' attention maps, as they reflect the underlying response to the character's final generated position \cite{wang2023high,wang2024primecomposer}. Therefore, we visualize the attention maps of their problematic results by rendering attention directly onto the images. We observe certain characters' attention misalignment, resulting in character duplication (see Fig. \ref{fig:visual_problem} (a) and (c)). Besides, excessive dispersion of attention also leads to character absence  (see Fig. \ref{fig:visual_problem} (b)). 
This observation underscores the insufficiency of these methods in guiding the model autonomously estimating optimal character-generated position.
% guiding the characters' attention toward their optimal generation areas during inference.
Further, we also assess specific methods \cite{UdiffText, TextDiffuser} quantitatively on two datasets to evaluate their proficiency in directing characters' attention toward their optimal generation regions throughout multiple global training steps.
Specifically, for each test sample, we extract potential position masks for all characters based on their cross-attention maps from the inference process. Then, we compute the mean Intersection over Union (mIoU) between these masks and ground truth character segmentation masks. 
Their inoptimal results depicted in Fig. \ref{fig:mIoU} further corroborate our hypothesis.

In summary, current methods encounter challenges in accurately rendering text within complex scenes due to their constrained representation domain and synthesizing characters in their ideal regions due to the lack of effective guidance for the model to estimate optimal character-generated position. 
To this end, this paper proposes \ourmethod for high-fidelity text synthesis. 
We reconstruct the diffusion training process, embedding more refined guidance: we expose and rectify the characters' attention to focus more precisely on their ideal generation areas, and enhance the learning of text regions and character representation by additional constraints.
However, this reconstruction inevitably introduces a complex hybrid optimization problem encompassing both discrete and continuous variables. Therefore, we design a heuristic alternate optimization strategy. Concurrently, we jointly train the text encoder and the generator, leveraging the diverse font styles within the training dataset to enrich character representation space. We demonstrate that this joint training can be seamlessly integrated into the heuristic alternate optimization process, facilitating a synergistic interplay between character representations learning and character attention re-estimation. Specifically, in each iteration, the pipeline begins by encoding the potential character-generated position information from cross-attention maps into latent character masks. These masks are subsequently employed to refine the representation of specific characters within the text encoder, facilitating the calibration of the characters' attention in the subsequent step. This iterative process contributes to both learning of better character representation and, as a by-product, explicit guidance for autonomously estimating character position.

Notably, the generator may initially struggle to direct attention to the desired generation region for each character, resulting in suboptimal latent masks and impacting training progress. Unlike existing methods \cite{TextDiffuser,UdiffText} that rigidly constrain character attention, we employ a balanced supervision strategy. Initially, we assist in calibrating the attention using character segmentation masks for a warm-up in the earlier stage. Once the model has preliminarily acquired the ability to estimate ideal generation positions, we remove this guidance, allowing the model to autonomous learning iteratively. 
This strategy strikes a balance between constraining the model and unleashing its flexibility in estimating optimal generation positions.
Fig. \ref{fig:mIoU} illustrates the efficacy of our method in continuously improving the generator's ability to estimate the optimal characters' position. 

Our contributions are:
(1) The proposed \ourmethod effectively alleviates the issues of character repetition, absence, and distortion encountered by existing methods.
(2) Our heuristic alternate optimization strategy, integrating the joint learning of the text encoder and U-Net, orchestrates a symbiotic relationship between learning character representations and re-estimating character attention.
(3) Our balanced supervision strategy strikes a balance between constraining the model and
unleashing its flexibility in estimating optimal generation positions.
(4) Both qualitative and quantitative results demonstrate the superiority of our method.

\begin{figure*}[ht]
\centering
\includegraphics[width=1\textwidth]{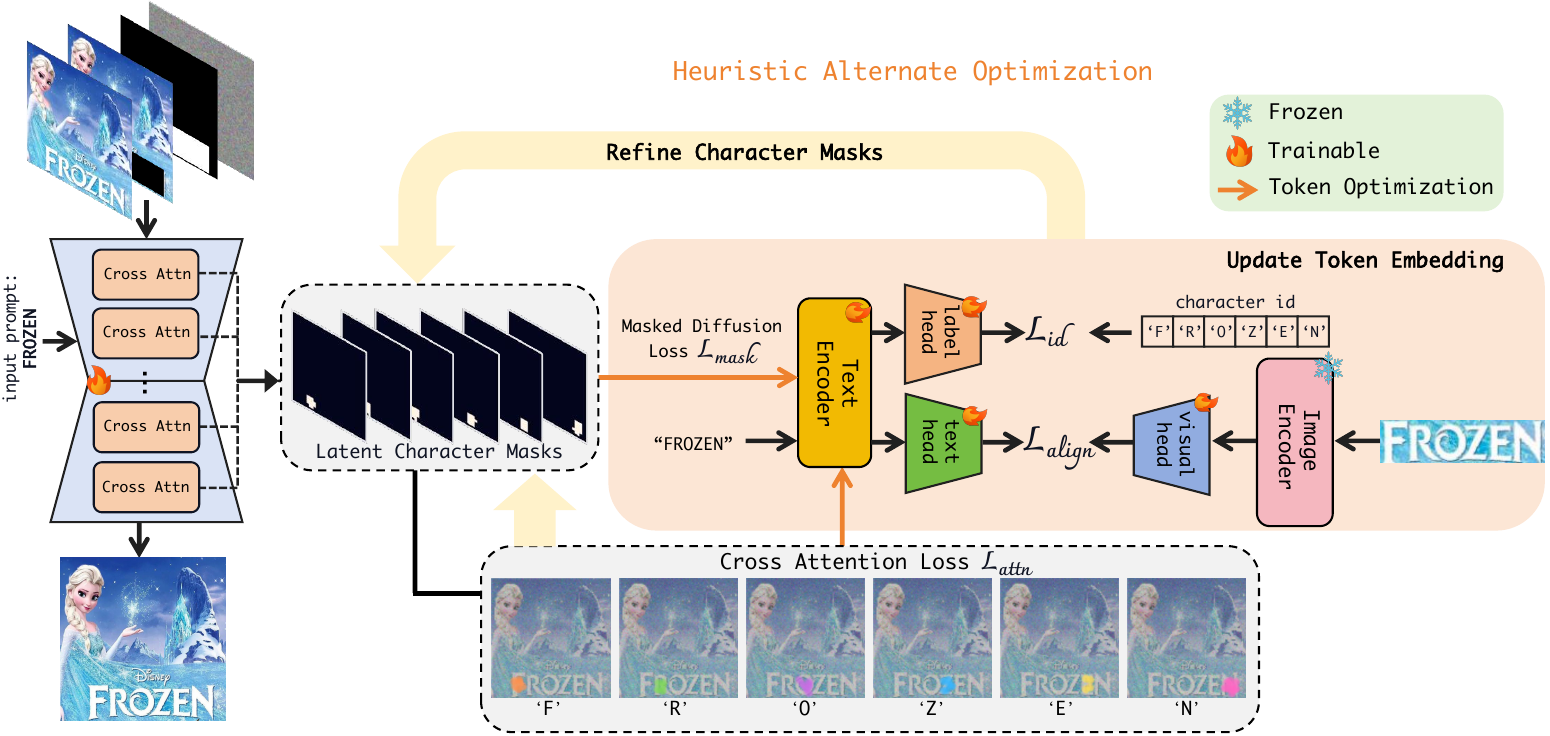}
\caption{An overview of proposed heuristic alternate optimization strategy.}
\label{fig:model}
\vspace{-0.4cm}
\end{figure*}

\section{Related Work}
% \subsection{Diffusion Models}
% Recent advances in diffusion models have achieved significant success in image synthesis and editing \cite{balaji2022ediff,li2023guiding,brack2023sega,betker2023improving,podell2023sdxl,saharia2022photorealistic}. 
% The leading diffusion model \cite{rombach2022high}, Stable diffusion, excels in effective text-conditional image generation. It achieves this by initially compressing images into a low-dimensional space using an autoencoder and subsequently leveraging text encodings with a cross-attention layer. This model is highly versatile and can be readily adapted for tasks such as image inpainting, text-guided image editing, etc. 
% Therefore, this work builds upon Stable Diffusion and utilizes relevant efficient sampling algorithms. \cite{karras2022elucidating,song2020score}.

% \subsection{Scene Text Synthesis}
Earlier GAN-based methods achieve scene text synthesis by transferring the text style in a reference image to the rendered target text image \cite{kong2022look,roy2020stefann,shimoda2021rendering,yang2020swaptext,wang2023gan}. Specifically, STEFANN \cite{roy2020stefann} employs a FANnet to edit individual characters and incorporates a placement algorithm to generate the desired word. SRNet \cite{wu2019editing} and MOSTEL \cite{MOSTEL}, on the other hand, split the task into two main parts: background inpainting and text style transfer. This approach allows for the end-to-end synthesizing of entire words. Despite their simplicity and effectiveness, they are limited in their ability to generate text in arbitrary styles and locations, often producing less natural-looking images.

To this end, several diffusion-based approaches have emerged to address these challenges. These approaches harness the robust capabilities of diffusion models to edit or generate scene text, thereby enhancing the quality and diversity of the generated content. 
DiffSTE \cite{DiffSTE} proposes a dual encoder structure, comprising a character text encoder and an instruction text encoder. These components are employed for instruction tuning, providing improved control over the backbone network. DiffUTE \cite{chen2024diffute} employs an OCR-based glyph encoder to extract glyph guidance from the rendered glyph image. 
% Similarly, GlyphDraw \cite{ma2023glyphdraw} integrates an additional image encoder and a fusion module to incorporate glyph conditions, enabling the fine-tuned model to generate images with coherent Chinese text. GlyphControl \cite{yang2024glyphcontrol} employs ControlNet \cite{zhang2023adding} for text image generation tasks, utilizing the rendered reference image as guidance for both position and glyph. 
TextDiffuser \cite{TextDiffuser} concatenates the segmentation mask as conditional input and employs character-aware loss to control the generated characters precisely. However, they still face challenges in generating accurate text within polystylistic images, primarily due to the insufficient conditional guidance provided by their character-unaware text encoder. 
Motivated by this, UDiffText \cite{UdiffText} replaces the original CLIP text encoder in Stable Diffusion with a character-level text encoder with a unitary font. It also employs local attention loss for position control, relying on ground truth character segmentation maps. 

However, these methods still encounter challenges in accurately rendering text within complex scenes and synthesizing characters in their ideal regions, attributed to their constrained character representation domain and lack of effective guidance for the model to autonomously estimate character-generated position, respectively.
% Therefore, this paper presents DreamText, featuring our novel alternate optimization strategy, which orchestrates a symbiotic relationship between the refinement of attention maps and the optimization of character representations via iterative optimization. Besides, our semi-supervised learning strategy for latent character masks enables fine-tuning on additional datasets lacking character-level segmentation masks, leading to more high-fidelity scene text synthesis.

\section{Method} \label{sec: method}
% In this section, we will first provide an overview of our framework and briefly introduce the Latent Diffusion Model. Then, We will describe the pipeline for acquiring latent character masks. Subsequently, we will delve into the details of our alternate optimization strategy. Finally, we will introduce our semi-supervised learning trick.

Given an input image $\textbf{\textit{z}}$, a text region specified by a binary mask $\textbf{\textit{B}}$ and a text condition $\textit{c}$, scene text synthesis is designed to render the given text onto the input image at the given text region. A natural idea for scene text synthesis is to formulate it into a latent diffusion process \cite{rombach2022high}, which is equivalent to solving the following optimization problem:
\begin{equation}\label{L_ldm}
\min_{(\theta, \vartheta)} \mathcal{L}_{LDM} \triangleq \mathbb{E}_{\textbf{\textit{z}}, c,\bm{\epsilon} \sim N(0,1), t}  \parallel \bm{\epsilon} - \epsilon_{\theta}(\textbf{\textit{z}}_{t}, t, \psi_{\vartheta}(c), \textbf{\textit{B}}) \parallel_{2}^{2},
\nonumber
\end{equation}
where  $\psi_{\vartheta}$ is the learnable character-level text encoder initialized by UDifftext \citep{UdiffText} and $\theta$ is the model parameter vector. However, we argue that such a straightforward formulation could be problematic due to the following reasons:
\begin{itemize}
    \item Without explicit guidance for rectifying characters' attention, character repetition and absence can inevitably occur in the rendered results as shown in Fig. \ref{fig:problem}.
    \item This formulation uniformly measures the mean distance between all pixels, without emphasizing the text regions. Consequently, text-unrelated regions can adversely affect the learning of the text encoder during backpropagation. The experimental analysis is provided in Sec. \ref{sec: ablation}.
    \item Due to the intricate interactions among the U-Net and text encoder, end-to-end training is particularly challenging as shown in Sec. \ref{sec: ablation}. Therefore, a more refined guidance for the training process is necessary.
\end{itemize}

% This work aims to propose a framework for high-fidelity scene text synthesis. The latent diffusion model \cite{rombach2022high} is utilized as our generator $\epsilon_\theta$. 
% Given an input image $\textbf{\textit{z}}$, a binary mask $\textbf{\textit{B}}$ of the text region, and text condition $\textit{c}$, the generator renders the specified text onto the input image at the positions indicated by \textbf{\textit{B}} and produces the resulting image as output. 
% When training the diffusion model based on its basic objective, the loss function is:
% \begin{equation}\label{L_ldm}
% \mathcal{L}_{LDM} = argmin_{(\theta, \vartheta)}\mathbb{E}_{\textbf{\textit{z}}, c,\bm{\epsilon} \sim N(0,1), t}  \parallel \bm{\epsilon} - \epsilon_{\theta}(\textbf{\textit{z}}_{t}, t, \psi_{\vartheta}(c), \textbf{\textit{B}}) \parallel_{2}^{2} 
% \end{equation}
% where $\psi$ is the learnable character-level text encoder initialized from \cite{UdiffText}. However, relying solely on this objective is problematic: (1) the lack of guidance for characters' synthesis position leads to deflected attention which is the main reason for character repetition and absence; (2) the text encoder struggles to learn fine-grained and robust character representations effectively because Equ. \ref{L_ldm}  measures the mean distance between all pixels, neglecting the critical need for precise character rendering; (3) the simultaneous learning of the text encoder and U-Net, without an effective strategy, hinders model convergence.

In this paper, we propose \ourmethod and develop a heuristic iterative optimization pipeline to enable efficient training. Our key idea is to explicitly model the potential character positions in attention maps by encoding them into latent character masks in each step. These masks are then used to enhance the learning of the text encoder through our carefully crafted losses, facilitating the calibration of character attention in the subsequent step. This heuristic iterative process allows our model to dynamically alternate between optimizing character embedding and re-estimating character generation positions. The details are presented in the following sections. 

\subsection{Latent Diffusion with Refined Guidance} \label{loss}
\begin{figure*}[htb]
\centering
\includegraphics[width=0.9\textwidth]{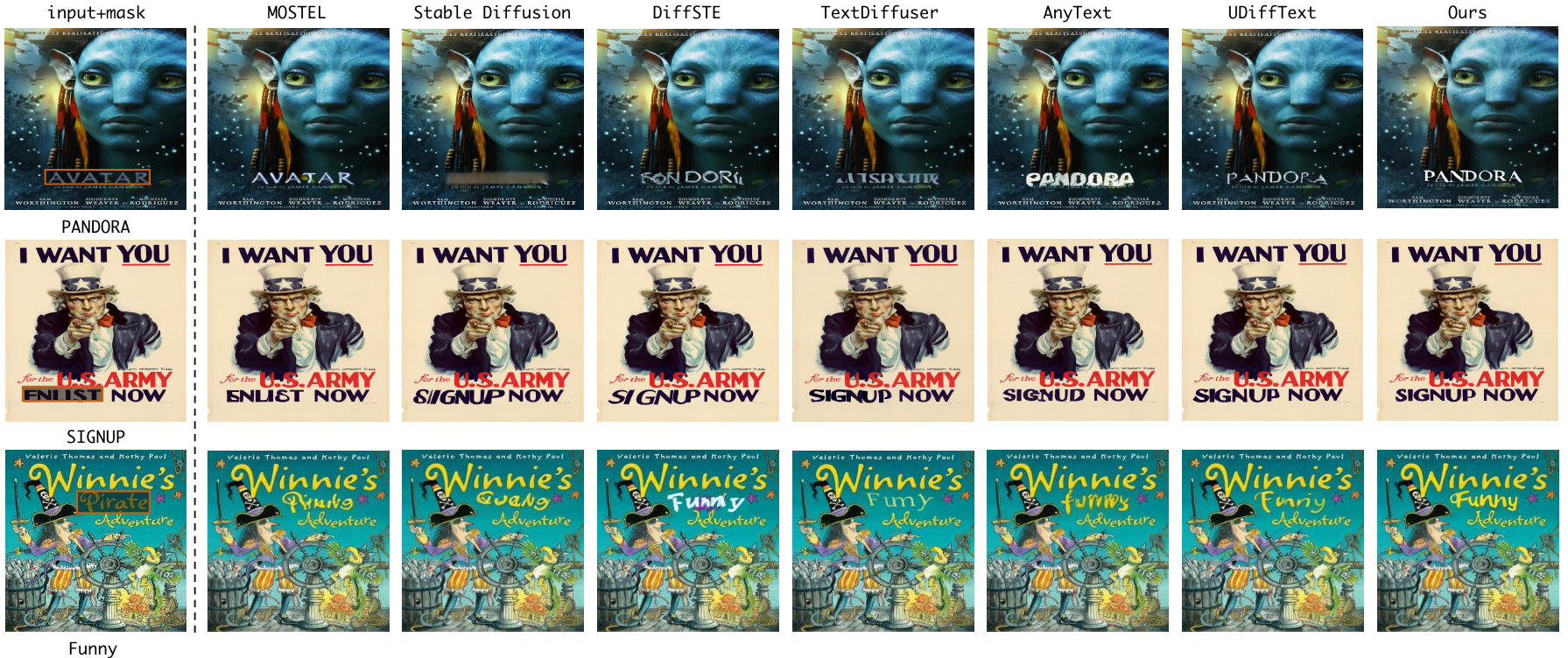}
\caption{Qualitative comparative results against state-of-the-art methods.}
\label{fig:compare}
\vspace{-0.4cm}
\end{figure*}

\subsubsection{Latent Character Mask} \label{sec:mask}
We utilize cross-attention maps to generate latent character masks, extracting characters' latent position information in the current step. Given latent image $\textbf{\textit{z}}_t$ and text embedding $\psi_{\vartheta}(\textit{c})$, the cross-attention map in layer \textit{l} is defined as follows:
\begin{align}
\textbf{\textit{Q}}_{l}=\textbf{\textit{z}}_t\textbf{\textit{W}}_{l}^{q},\quad\textbf{\textit{K}}_{l}=\psi_{\vartheta}(\textit{c})\textbf{\textit{W}}_{l}^{k},  \\
    \textbf{\textit{A}}_l=\mathrm{softmax}\left(\frac{\textbf{\textit{Q}}_l\textbf{\textit{K}}_{l}^T}{\sqrt{d}}\right), 
\end{align}
where $\textbf{\textit{W}}_{l}^{q}$ and $\textbf{\textit{W}}_{l}^{k}$ are learnable parameters and \textit{d} is the embedding dimension. The attention maps $\textbf{\textit{A}}_l$ is reshaped to $N \times H \times W$, where \textit{N} represents the number of text tokens such that each slice $\textbf{\textit{A}}_{l}^{i} \in \mathbb{R}^{H \times W}$ represents the region attended by the $i$-th token. 
These attention maps are then averaged across all layers of the U-Net to get the mean response $\bar{\textbf{\textit{A}}} = \frac{1}{L}\sum_{l=1}^{L} \textbf{\textit{A}}_l$ for the regions attended by character tokens. 

To obtain the latent character masks $\textbf{\textit{M}} \in \mathbb{R}^{N \times H \times W}$, we first apply Gaussian blur to the attention map to execute low-pass filtering $\mathrm{blur}(\bar{\textbf{\textit{A}}})$. This step helps mitigate excessive variance in the attended regions, ensuring a more uniform distribution of attention across relevant character regions.
Subsequently, we employ a straightforward thresholding process to convert the blurred attention map into binary character masks. 
This process is encapsulated by the function $f(\cdot)$, which applies a threshold and assigns a value of 1 to pixel values exceeding the threshold, and 0 otherwise.  The threshold is calculated as the mean value plus twice the variance of averaged attention maps in this work. That is:
\begin{align} 
    f(\textbf{\textit{X}}) = \begin{cases}
        1,\mbox{ if } x_{i,j}> \mbox{mean}(\textbf{\textit{X}}) + 2 \mbox{std}(\textbf{\textit{X}})\\
        0, \mbox{otherwise}
    \end{cases}.
\end{align}
Therefore, the equation for obtaining $\textbf{\textit{M}}$ can be expressed as:
\begin{equation}\label{get_mask}
    \textbf{\textit{M}}=f(\mathrm{blur}(\bar{\textbf{\textit{A}}})).
\end{equation}

Based on the latent character masks $\textbf{\textit{M}}$, we employ several loss functions to optimize both the text encoder and U-Net iteratively. We will delve into these in the following sections.

\subsubsection{Masked Diffusion Loss} \label{loss_masked}
We first extend the Eq. \ref{L_ldm} to highlight all desired character tokens in the current step. Specifically, within a text \textit{c} containing \textit{k} tokens of interest, the diffusion loss of the corresponding pixels obtained from each token's latent character mask is applied with an additional weighting factor $\gamma$. Formally, considering $\textbf{\textit{M}}_{k}=\bigvee_{i=1}^{k}\textbf{\textit{M}}_{i}$ to the union of the pixels of \textit{k} characters, the masked diffusion loss is formulated as,
\begin{equation}
\mathcal{L}_{mask} = \mathbb{E}_{\textbf{\textit{z}}, c,\bm{\epsilon} \sim N(0,1), t}  \parallel (1+\gamma\textbf{\textit{M}}_k)(\bm{\epsilon} - \epsilon_{\theta}(\textbf{\textit{z}}_{t}, t, \psi_{\vartheta}(c))) \parallel_{2}^{2} . \nonumber
\end{equation}
We omit binary mask \textbf{\textit{B}} here for the simplicity.

\subsubsection{Cross Attention Loss} \label{loss_cross}
The masked diffusion loss is designed to focus on all desired concept tokens within a prompt. Additionally, to ensure that each token encodes information specific to its corresponding synthesis position, we incorporate the cross-attention loss \cite{avrahami2023break}. This loss encourages tokens to attend exclusively to their corresponding target regions, 
\begin{equation}
    \mathcal{L}_{attn}=\mathbb{E}_{\textbf{\textit{z}}, c, t} \parallel C_{attn}(\textbf{\textit{z}}_t,\psi_{\vartheta}(c)_i) -\textbf{\textit{M}}_i \parallel_2^2.
\end{equation}
Here, $C_{attn}(\textbf{\textit{z}}_t,\psi_{\vartheta}(c)_i)$ represents the cross-attention map between the visual representation $\textbf{\textit{z}}_{t}$ and the token $\psi_{\vartheta}(c)_i$, while $\textbf{\textit{M}}_{i}$ denotes the latent character mask of $\psi_{\vartheta}(c)_i$.

However, the losses mentioned above alone are inadequate for achieving optimal character representation. This is primarily due to the inherent noise in the latent character masks, which frequently results in under- or over-segmentation of the ideal character regions during training. To mitigate the influence of noise and prevent skewed learning of character representation, we introduce additional losses \cite{UdiffText} as shown below. 

\subsubsection{Cross-modal Aligned Loss} \label{loss_align}
To obtain accurate and robust embeddings, we introduce an image encoder $\xi$, text head $\textit{H}_t$, and visual head $\textit{H}_v$ to align cross-modal character features. Specifically, we compute the cosine similarity objective between visual and text representations to maximize the alignment between two modalities:
\begin{equation}
\mathcal{L}_{align}=\frac{\langle \textit{H}_t(\textbf{\textit{y}}),\textit{H}_v(\xi(\textbf{\textit{I}}))\rangle}{\|\textit{H}_t(\textbf{\textit{y}})\|_2\cdot\|\textit{H}_v(\xi(\textbf{\textit{I}}))\|_2}.
\end{equation}
where \textbf{\textit{I}} represents the text image segmented via the bounding box from datasets and $\langle, \rangle$ presents the inner product of two vectors. Note that to alleviate noise stemming from background and color variations, we preprocess the text image by converting it to grayscale. 
% This alignment encourages the model to learn discriminative embeddings that capture both visual and textual characteristics of characters, facilitating more effective representation learning.

Through training, the text head learns to map the whole information of multiple characters into the text feature space. Similarly, the image head maps the overall visual text information into the visual feature space. This effectively ensures the overall textual and visual representations correspond, capturing high-level cross-modal consistency.

\subsubsection{Character Id Loss} \label{loss_id}
Additionally, we introduce character Id loss to ensure that the learned embeddings are highly distinguishable. To be precise, a multi-label classification head $\textit{H}_l$ is introduced to predict character indices from text embeddings $\textit{\textbf{y}}$. Then, the cross-entropy objective is formulated,  
\begin{equation}
\mathcal{L}_{id} = - \sum_{i=1}^{N} \sum_{j=1}^{K} \textit{\textbf{l}}_{{i,j}} \log(\textit{H}_l(\textit{\textbf{y}})_j), \label{id_loss}
\end{equation}
where \textit{N} and \textit{K} represent the number of characters and the possible indices respectively while \textbf{\textit{l}} denotes the ground truth labels.
This loss aggregates over all characters in the target text, ensuring the text encoder produces distinguishable embeddings for each character.

The complete objective of our training strategy can be expressed as
\begin{equation}
\mathcal{L}=\mathcal{L}_{mask}+\alpha\mathcal{L}_{attn}+\beta(\mathcal{L}_{align}+\mathcal{L}_{id}) .
\end{equation}

\begin{table*}[ht]  \centering
	\scriptsize
        \setlength{\tabcolsep}{5pt}
	\caption{Quantitative comparison against baselines.}
 \vspace{-0.2cm}
	\label{table1}
	\begin{tabular}{c|cccc|cccc|cc}
		\bottomrule
		\multirow{2}{*}{\textbf{Methods}} & \multicolumn{4}{c|}{\textit{\textbf{SeqAcc-Recon}}} & \multicolumn{4}{c|}{\textit{\textbf{SeqAcc-Editing }}} \\
		& ICDAR13(8ch)      & ICDAR13     &TextSeg & LAION-OCR     & ICDAR13(8ch)      & ICDAR13     &TextSeg & LAION-OCR   &FID &LPIPS  \\
		\bottomrule

            CVPR'22 SD-Inpainting \cite{rombach2022high}                 & 0.32     & 0.29& 0.11& 0.15& 0.08        & 0.07   &0.04      & 0.05     & 26.78& 0.0696   \\
  	     arXiv'23 DiffSTE \cite{DiffSTE}      & 0.45       & 0.37& 0.50& 0.41& 0.34      & 0.29    & 0.47     & 0.27 & 51.67& 0.1050   \\
            AAAI'23 MOSTEL \cite{MOSTEL}                   & 0.75      & 0.68& 0.64& 0.71 & 0.35       & 0.28   &0.25      &0.44    & 25.09& 0.0605     \\

		NIPS'23 TextDiffuser \cite{TextDiffuser}  & 0.87       & 0.81& 0.68& 0.80& 0.82      & 0.75   &0.66      & 0.64    & 32.25& 0.0834 \\
  		ICLR'24 AnyText   \cite{tuo2023anytext}        & 0.89      & 0.87 & 0.81& 0.86& 0.81       & 0.79      &0.80   &0.72  & 22.73& 0.0651   \\
		ECCV'24 UDiffText   \cite{UdiffText}        & 0.94      & 0.91& 0.93& 0.90& 0.84       & 0.83      &0.84   &0.78  & 15.79& 0.0564   \\

            \bottomrule
		\textbf{DreamText}              & \textbf{0.95} & \textbf{0.94}& \textbf{0.96}& \textbf{0.93}& \textbf{0.87}   &\textbf{0.89}        & \textbf{0.91}         & \textbf{0.88}  & \textbf{12.13}& \textbf{0.0328}  \\
		\bottomrule
	\end{tabular} \\

	\label{table:compare}
\vspace{-0.3cm}
\end{table*}

\subsection{Optimization Strategy}
\subsubsection{Heuristic Alternate Optimization} Our objective involves numerous discrete variables, rendering it non-differentiable and making vanilla SGD invalid. Thus, we employ a heuristic alternate optimization strategy, utilizing latent character masks that encapsulate potential synthesis position information, to facilitate a symbiotic relationship between text encoder and U-Net. Specifically, we perform an alternating update between the tokens given the latent character masks and the masks themselves. During optimization, the masks are computed by Eq. \ref{get_mask}, while for other parameters, we fix the masks and calculate gradients accordingly. Using the losses above, we first optimize the representation of specific characters given masks in each iteration, which, in turn, enables the generator to rectify the character’s attention in subsequent steps.
% In the subsequent step, the optimized tokens aid in calibrating character attention, thereby refining the latent character masks. 
This process allows model to dynamically alternate between optimizing character embeddings and re-estimating character masks.

\subsubsection{Balanced Supervision on Character Attention} The initial training phase may challenge the generator in directing attention to the intended generation region for each character. Unlike existing methods that excessively constrain character attention which inevitably limits the model's flexibility, we adopt a balanced supervision. Initially, we guide attention calibration by cross-entropy objective between latent character masks and character segmentation masks from datasets similar to Eq. \ref{id_loss} for a warm-up in the earlier stage. Once the model preliminarily acquires the capability to estimate ideal generation positions (approximately 25k steps as shown in Fig. \ref{fig:mIoU}), we cease guidance, enabling it to engage in autonomous iterative learning. This strategy balances constraining the model and unleashing its flexibility in estimating characters' optimal generation positions.
% This strategy further elevates performance, expanding the model's capacity to adapt to diverse datasets that lack character segmentation masks.

\section{Experiments}
\subsection{Implementation Details}
% \subsubsection{Datasets}
% Several datasets are employed to evaluate the competitive methods: SynthText \cite{gupta2016synthetic}, LAION-OCR \cite{TextDiffuser}, ICDAR13 \cite{karatzas2013icdar} and TextSeg \cite{xu2021rethinking}. We provide a detailed explanation in the appendix.
% \input{figs/compare1}

\subsubsection{Datasets}
Several datasets are used in our experiments:
(1) \textbf{SynthText} \cite{gupta2016synthetic} comprises 800,000 images containing around 8 million synthetic word instances. Each text instance is annotated with its corresponding text string, along with word-level and character-level bounding boxes.
(2) \textbf{LAION-OCR} \cite{TextDiffuser} contains 9,194,613 filtered high-quality text images including advertisements, notes, posters, covers, memes, logos, etc.
(3) \textbf{ICDAR13} \cite{karatzas2013icdar} is widely recognized as the benchmark dataset for evaluating near-horizontal text detection, which consists of 233 test images used for evaluation purposes.
(4) \textbf{TextSeg} \cite{xu2021rethinking} comprises 4,024 real-world text images sourced from various sources such as posters, greeting cards, book covers, logos, road signs, billboards, digital designs, handwritten notes, and more. Note that only SynthText and LAION-OCR datasets provide the character segmentation maps. We train our model using the training subsets of these datasets and randomly select 100 images from test subsets for testing.

\subsubsection{Training Configurations}
We train our model based on the pre-trained checkpoint of SD-v2.0 inpainting version and text encoder in \cite{UdiffText}. The pre-trained Vision Transformer \cite{atienza2021vision} is used as our image encoder. The model is fine-tuned using 4 NVIDIA A100 GPUs on LAION-OCR for 200k steps, SynthText for 150k steps, TextSeg for 50k steps, and ICDAR13 for an additional 10k steps. We set $\alpha$ to 0.01, and $\beta$ to 0.001. Additionally, we utilize a batch size of 16 and a learning rate of 5 $\times$ $10^{-5}$.
The inference time of our \ourmethod is 8.5 seconds only.

\begin{figure*}[ht]
\centering
\includegraphics[width=0.9\textwidth]{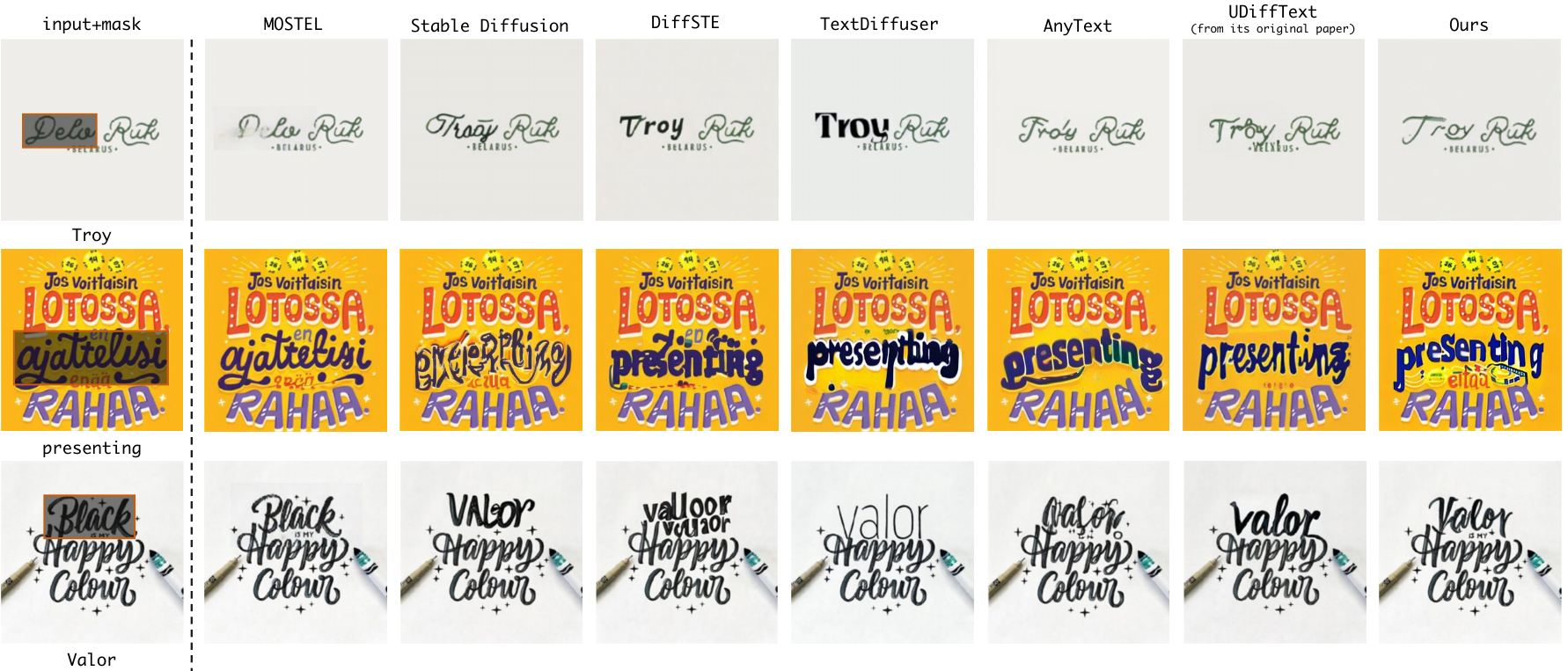}
\caption{Qualitative comparative results against state-of-the-art methods.}
\label{fig:orig_compare}
\vspace{-0.4cm}
\end{figure*}
\subsubsection{Evaluation}
\textbf{Text accuracy}: We assess all methods on two tasks: scene text reconstruction and scene text editing \cite{UdiffText}. For the scene text reconstruction task, we use the models to reconstruct the text image using the provided ground truth text label and binary mask. In the scene text editing task, we replace the original text in each image with a random word and evaluate the models by generating images containing the edited text. We use an off-the-shelf scene text recognition (STR) model \cite{bautista2022scene} to identify the rendered text and then evaluate word-level correctness using sequence accuracy (SeqAcc) by comparing the STR result with the ground truth. 

\noindent\textbf{Image quality}: 
we utilize Fréchet Inception Distance (FID) \cite{heusel2017gans} to quantify the distance between the text images in the dataset and the generated images. Additionally, the Learned Perceptual Image Patch Similarity (LPIPS) \cite{zhang2018unreasonable} is utilized as an additional metric.

\begin{table}[t]  
	\small
 \centering
        \setlength{\tabcolsep}{11pt}
	% \caption{Ablation study results on different settings.}
        
        \caption{Accumulated results of losses.}
        % \vspace{-0.3cm}
	\begin{tabular}{c|cc|cc}
		\bottomrule
		\multirow{2}{*}{\textbf{Setting}} & \multicolumn{2}{c|}{\textit{\textbf{Average SeqAcc}}} & \\
            & Recon & Editing &FID &LPIPS\\
		\bottomrule
            Base                  & 0.218      &0.060   & 26.78   & 0.0696\\
		+$\mathcal{L}_{mask}$         &0.425      & 0.259     & 23.21  & 0.0528\\
	     +$\mathcal{L}_{attn}$         & 0.698    & 0.532 & 19.72  &0.0483\\
		+$\mathcal{L}_{align}$      &0.884      & 0.801    & 15.41 & 0.0392\\
		+$\mathcal{L}_{id}$           &\textbf{0.940}   &\textbf{0.887}  & \textbf{12.13}  & \textbf{0.0328} \\
		\bottomrule
	\end{tabular} \\
        \label{table:ablation_setting}
	% \label{table:ablation}
\vspace{-0.3cm}
\end{table}

\subsection{Results}
\textbf{Quantitatively}: In Tab. \ref{fig:compare}, we present a quantitative analysis comparing \ourmethod against baseline methods. Our method demonstrates a significant advantage across all quantitative metrics, highlighting the exceptional visual quality of the scene text images it generates. Notably, when compared to the previous state-of-the-art model UDiffText, \ourmethod outperforms it by 3.66 in terms of FID. While UDiffText and AnyText excel in image quality compared to other methods, except \ourmethod, they tend to perform less satisfactorily in terms of sequence accuracy, especially in text editing. This discrepancy may be attributed to its restricted representation domain and deflected attention, resulting in less accurate text rendering within complex scenes.

\noindent\textbf{Qualitatively}: As depicted in Fig. \ref{fig:compare}, our method outperforms other baselines in synthesizing more coherent and accurate text within polystylistic scenes. For example, in Fig. \ref{fig:compare} (3rd row), MOSTEL and Stable Diffusion fail to generate the text, while TextDiffuser and UDiffText encounter the issue of character absence. Similarly, in Fig. \ref{fig:compare} (1st row), all baselines fail to generate the text on the provided poster. Moreover, we further compare our method with the latest approach, UDiffText, using the samples provided in its original paper for additional qualitative evaluation. The visualization results are presented in Fig. \ref{fig:orig_compare}, offering additional evidence of our method's superiority in generating high-fidelity text.
\begin{table}[]  
	\small
        \setlength{\tabcolsep}{6pt}
	
        \caption{Hyperparameter analysis.}
        % \vspace{-0.3cm}
	\begin{tabular}{c|cc|c|cc}
		\bottomrule
		\multirow{1}{*}{$\alpha$} & \multicolumn{2}{c|}{\textit{\textbf{Average SeqAcc}}} & \multirow{1}{*}{$\beta$} & \multicolumn{2}{c}{\textit{\textbf{Average SeqAcc}}} \\
            ($\beta=0$) & Recon & Editing & ($\alpha$=0.01) & Recon & Editing \\
		\bottomrule
		1         &0.672      & 0.558   & 1         &0.864      & 0.766   \\
	    0.1         & 0.724    & 0.615  &0.1         & 0.896    & 0.817  \\
		\textbf{0.01}      &0.748      & \textbf{0.623} &0.01 &0.921   &0.864   \\
		0.001          &\textbf{0.753}   &0.604  &\textbf{0.001}        &\textbf{0.940}      & \textbf{0.887}  \\
		\bottomrule
	\end{tabular} \\
        \label{table:ablation_weight}
	% \label{table:ablation_weight}
        \vspace{-0.3cm}
\end{table}

\subsection{Ablation Studies and Effectiveness Analysis} \label{sec: ablation}
\textbf{Ablation on Losses}: We employ SD-v2.0 inpainting version as the base method. As shown in Tab. \ref{table:ablation_setting}, we observe a significant drop in performance across all metrics when these losses are removed.
Additionally, these results highlight that the masked diffusion loss and cross-attention loss alone are insufficient for achieving satisfactory character representation. This may be attributed to the inherent noise in the latent character masks. To sum up, these ablation results underscore the effectiveness of each objective within our heuristic alternate optimization strategy in improving various aspects of synthesized scene texts.

\noindent\textbf{Choice of $\alpha$ and $\beta$}: In our experiments, we vary the values of $\alpha$ and $\beta$ within the range [0.001, 0.01, 0.1, 1]. The results are presented in Tab. \ref{table:ablation_weight}. Initially, we focus on exploring the impact of $\alpha$ while setting $\beta$ = 0. We find that the optimal performance in text reconstruction occurs when $\alpha$ = 0.001. However, this choice yields poor results in editing tasks. Consequently, we opt for a more balanced approach, selecting $\alpha$ = 0.01. Subsequently, with $\alpha$ fixed at 0.01, we investigate the effect of $\beta$. We find that the optimal performance is achieved when $\beta$ = 0.001.
\begin{figure}[t]
\centering
\includegraphics[width=1\linewidth]{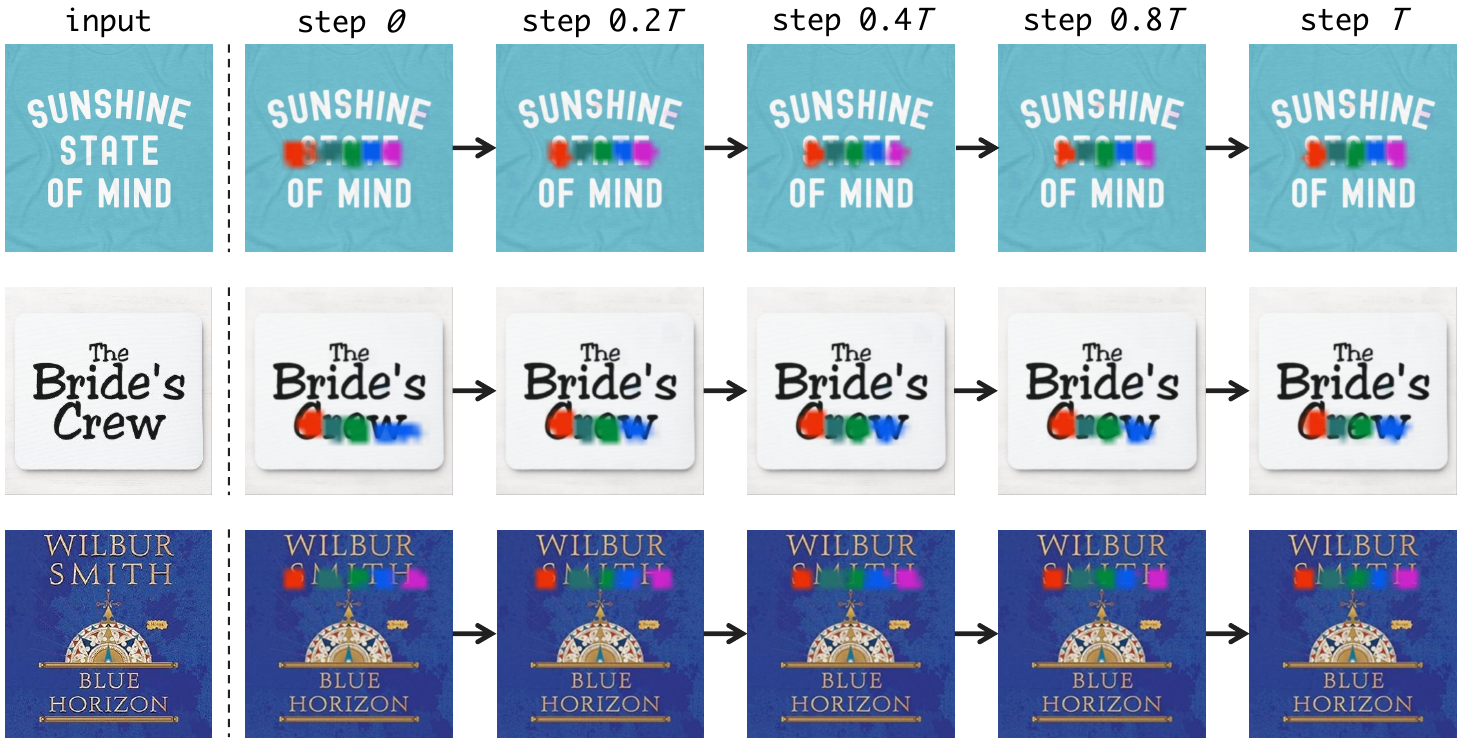}
\caption{The visualized attention results of all characters across several steps during training.}
\label{fig:visual_mid}
\vspace{-0.3cm}
\end{figure}

\noindent\textbf{Effectiveness of Heuristic Alternate Optimization}: We analyze the efficacy of our strategy by visualizing the evolution of character attention maps across several training steps. As illustrated in Fig. \ref{fig:visual_mid}, the model exhibits deflected attention initially but progressively corrects it in subsequent training steps. Ultimately, attention becomes concentrated on the desired positions for all characters, indicating the effectiveness of our approach. Besides, Fig. \ref{fig:mIoU} also illustrates the efficacy of this strategy in continuously improving the generator’s ability to estimate the optimal characters’ position by autonomously rectifying the attention during training.

\begin{table}[!t]  
	\small
        \centering
        \setlength{\tabcolsep}{10.5pt}
        \caption{Ablation study results on warm-up steps.}
	\begin{tabular}{c|cc|cc}
		\bottomrule
		 & \multicolumn{2}{c|}{\textit{\textbf{Average SeqAcc}}}& \\
            & Recon
$\uparrow$& Editing $\uparrow$&FID $\downarrow$&mIoU $\uparrow$\\
		\bottomrule
            15k                  & 0.884      &0.852    & 13.82   & 0.681\\
		  20k      &0.913      & 0.873     & 13.38 &  0.692\\
    	  \textbf{25k}         & \textbf{0.940}    & 0.887 & \textbf{12.13} & \textbf{0.722}\\

          30k         & 0.921    & \textbf{0.891} & 13.24 & 0.703\\
		% +$\mathcal{L}_{align}$      &0.873      & 0.712    & 26.53& \\
		% +$\mathcal{L}_{id}$           &\textbf{0.940}   &\textbf{0.847}  & \textbf{15.79} &  \\
		\bottomrule
	\end{tabular} \\
	\label{table:warm_up}
\vspace{-0.4cm}
\end{table}
\noindent\textbf{Choice of Warm-up Steps}:
We conduct an ablation experiment to analyze the impact of different warm-up steps on model performance. The results, presented in Tab. \ref{table:warm_up}, demonstrate how varying the number of warm-up steps affects the performance.
As the number of warm-up steps increases from 15k to 25k, we observe consistent improvements across all metrics.  However, at 30k steps, while the editing accuracy slightly improves, the FID score worsens, suggesting a potential trade-off between image fidelity and editing capability.
Overall, these findings highlight that 25k warm-up steps strike the best balance between reconstruction accuracy, editing performance, and image quality, making it the optimal choice for our method.

\section{Conclusion}
This paper presents \ourmethod, a novel approach for high-fidelity scene text synthesis.
We reconstruct the diffusion training process to expose and rectify the model's attention at the character level while strengthening its learning of text regions. This inevitably involves a hybrid optimization challenge, combining discrete and continuous variables, which we effectively address through our heuristic alternate optimization strategy. Meanwhile, we jointly train the text encoder and generator to comprehensively learn and utilize diverse fonts present in the training dataset, fostering a synergistic relationship between learning character embeddings and re-estimating character attention.
Both qualitative and quantitative results highlight the superiority of our method.

% featuring our heuristic alternate optimization strategy. This strategy cultivates a synergistic interplay between the text encoder and U-net, mitigating instability caused by their mutual influence during training. As a result, DreamText effectively alleviates issues like character repetition, absence, and distortion found in existing methods. Furthermore, our warm-up learning strategy for refining latent masks enables DreamText to adapt to datasets lacking character-level segmentation masks. Both qualitative and quantitative evaluations validate the superior performance of our method compared to the state of the art. 

\section{Acknowledgments}
This work was supported by the National Nature Science Foundation of China (62472097), AI for Science Foundation of Fudan University (FudanX24AI028), and Shanghai Innovation Institute. The computations in this research were performed on the CFFF platform of Fudan University. Cheng Jin is the corresponding author of this work.
{
    \small
    \bibliographystyle{ieeenat_fullname}
    \bibliography{11_references}

\begin{thebibliography}{29}
\providecommand{\natexlab}[1]{#1}
\providecommand{\url}[1]{\texttt{#1}}
\expandafter\ifx\csname urlstyle\endcsname\relax
  \providecommand{\doi}[1]{doi: #1}\else
  \providecommand{\doi}{doi: \begingroup \urlstyle{rm}\Url}\fi

\bibitem[Atienza(2021)]{atienza2021vision}
Rowel Atienza.
\newblock Vision transformer for fast and efficient scene text recognition.
\newblock In \emph{ICDAR}, pages 319--334, 2021.

\bibitem[Avrahami et~al.(2023)Avrahami, Aberman, Fried, Cohen-Or, and Lischinski]{avrahami2023break}
Omri Avrahami, Kfir Aberman, Ohad Fried, Daniel Cohen-Or, and Dani Lischinski.
\newblock Break-a-scene: Extracting multiple concepts from a single image.
\newblock In \emph{SIGGRAPH Asia}, pages 1--12, 2023.

\bibitem[Bautista and Atienza(2022)]{bautista2022scene}
Darwin Bautista and Rowel Atienza.
\newblock Scene text recognition with permuted autoregressive sequence models.
\newblock In \emph{ECCV}, pages 178--196, 2022.

\bibitem[Chen et~al.(2024)Chen, Xu, Gu, Li, Meng, Zhu, Wang, et~al.]{chen2024diffute}
Haoxing Chen, Zhuoer Xu, Zhangxuan Gu, Yaohui Li, Changhua Meng, Huijia Zhu, Weiqiang Wang, et~al.
\newblock Diffute: Universal text editing diffusion model.
\newblock In \emph{NeurIPS}, 2024.

\bibitem[Chen et~al.(2023)Chen, Huang, Lv, Cui, Chen, and Wei]{TextDiffuser}
Jingye Chen, Yupan Huang, Tengchao Lv, Lei Cui, Qifeng Chen, and Furu Wei.
\newblock Textdiffuser: Diffusion models as text painters.
\newblock In \emph{NeurIPS}, 2023.

\bibitem[Gupta et~al.(2016)Gupta, Vedaldi, and Zisserman]{gupta2016synthetic}
Ankush Gupta, Andrea Vedaldi, and Andrew Zisserman.
\newblock Synthetic data for text localisation in natural images.
\newblock In \emph{CVPR}, pages 2315--2324, 2016.

\bibitem[Heusel et~al.(2017)Heusel, Ramsauer, Unterthiner, Nessler, and Hochreiter]{heusel2017gans}
Martin Heusel, Hubert Ramsauer, Thomas Unterthiner, Bernhard Nessler, and Sepp Hochreiter.
\newblock Gans trained by a two time-scale update rule converge to a local nash equilibrium.
\newblock In \emph{NeurIPS}, 2017.

\bibitem[Ji et~al.(2023)Ji, Zhang, Wang, Hou, Zhang, Price, and Chang]{DiffSTE}
Jiabao Ji, Guanhua Zhang, Zhaowen Wang, Bairu Hou, Zhifei Zhang, Brian Price, and Shiyu Chang.
\newblock Improving diffusion models for scene text editing with dual encoders.
\newblock \emph{arXiv preprint arXiv:2304.05568}, 2023.

\bibitem[Karatzas et~al.(2013)Karatzas, Shafait, Uchida, Iwamura, i~Bigorda, Mestre, Mas, Mota, Almazan, and De~Las~Heras]{karatzas2013icdar}
Dimosthenis Karatzas, Faisal Shafait, Seiichi Uchida, Masakazu Iwamura, Lluis~Gomez i Bigorda, Sergi~Robles Mestre, Joan Mas, David~Fernandez Mota, Jon~Almazan Almazan, and Lluis~Pere De~Las~Heras.
\newblock Icdar 2013 robust reading competition.
\newblock In \emph{ICDAR}, pages 1484--1493, 2013.

\bibitem[Kong et~al.(2022)Kong, Luo, Ma, Zhu, Zhu, Yuan, and Jin]{kong2022look}
Yuxin Kong, Canjie Luo, Weihong Ma, Qiyuan Zhu, Shenggao Zhu, Nicholas Yuan, and Lianwen Jin.
\newblock Look closer to supervise better: One-shot font generation via component-based discriminator.
\newblock In \emph{CVPR}, pages 13482--13491, 2022.

\bibitem[Lakhanpal et~al.(2024)Lakhanpal, Chopra, Jain, Chadha, and Luo]{lakhanpal2024refining}
Sanyam Lakhanpal, Shivang Chopra, Vinija Jain, Aman Chadha, and Man Luo.
\newblock Refining text-to-image generation: Towards accurate training-free glyph-enhanced image generation.
\newblock \emph{arXiv preprint arXiv:2403.16422}, 2024.

\bibitem[Liu et~al.(2024)Liu, Liang, Liang, Luo, Li, Huang, and Yuan]{liu2024glyph}
Zeyu Liu, Weicong Liang, Zhanhao Liang, Chong Luo, Ji Li, Gao Huang, and Yuhui Yuan.
\newblock Glyph-byt5: A customized text encoder for accurate visual text rendering.
\newblock \emph{arXiv preprint arXiv:2403.09622}, 2024.

\bibitem[Park et~al.(2021)Park, Chun, Cha, Lee, and Shim]{park2021multiple}
Song Park, Sanghyuk Chun, Junbum Cha, Bado Lee, and Hyunjung Shim.
\newblock Multiple heads are better than one: Few-shot font generation with multiple localized experts.
\newblock In \emph{ICCV}, pages 13900--13909, 2021.

\bibitem[Qu et~al.(2023)Qu, Tan, Xie, Xu, Wang, and Zhang]{MOSTEL}
Yadong Qu, Qingfeng Tan, Hongtao Xie, Jianjun Xu, Yuxin Wang, and Yongdong Zhang.
\newblock Exploring stroke-level modifications for scene text editing.
\newblock In \emph{AAAI}, pages 2119--2127, 2023.

\bibitem[Rombach et~al.(2022)Rombach, Blattmann, Lorenz, Esser, and Ommer]{rombach2022high}
Robin Rombach, Andreas Blattmann, Dominik Lorenz, Patrick Esser, and Bj{\"o}rn Ommer.
\newblock High-resolution image synthesis with latent diffusion models.
\newblock In \emph{CVPR}, pages 10684--10695, 2022.

\bibitem[Roy et~al.(2020)Roy, Bhattacharya, Ghosh, and Pal]{roy2020stefann}
Prasun Roy, Saumik Bhattacharya, Subhankar Ghosh, and Umapada Pal.
\newblock Stefann: scene text editor using font adaptive neural network.
\newblock In \emph{CVPR}, pages 13228--13237, 2020.

\bibitem[Saharia et~al.(2022)Saharia, Chan, Saxena, Li, Whang, Denton, Ghasemipour, Gontijo~Lopes, Karagol~Ayan, Salimans, et~al.]{saharia2022photorealistic}
Chitwan Saharia, William Chan, Saurabh Saxena, Lala Li, Jay Whang, Emily~L Denton, Kamyar Ghasemipour, Raphael Gontijo~Lopes, Burcu Karagol~Ayan, Tim Salimans, et~al.
\newblock Photorealistic text-to-image diffusion models with deep language understanding.
\newblock In \emph{NeurIPS}, pages 36479--36494, 2022.

\bibitem[Shimoda et~al.(2021)Shimoda, Haraguchi, Uchida, and Yamaguchi]{shimoda2021rendering}
Wataru Shimoda, Daichi Haraguchi, Seiichi Uchida, and Kota Yamaguchi.
\newblock De-rendering stylized texts.
\newblock In \emph{ICCV}, pages 1076--1085, 2021.

\bibitem[Tuo et~al.(2024)Tuo, Xiang, He, Geng, and Xie]{tuo2023anytext}
Yuxiang Tuo, Wangmeng Xiang, Jun-Yan He, Yifeng Geng, and Xuansong Xie.
\newblock Anytext: Multilingual visual text generation and editing.
\newblock \emph{ICLR}, 2024.

\bibitem[Wang et~al.(2023{\natexlab{a}})Wang, Feng, Wu, Xu, and Zheng]{wang2023gan}
Yibin Wang, Yuchao Feng, Jie Wu, Honghui Xu, and Jianwei Zheng.
\newblock Ca-gan: Object placement via coalescing attention based generative adversarial network.
\newblock In \emph{ICME}, pages 2375--2380. IEEE, 2023{\natexlab{a}}.

\bibitem[Wang et~al.(2023{\natexlab{b}})Wang, Xu, Zhou, Zhang, and Jin]{wang2023enhancing}
Yibin Wang, Honghui Xu, Changhai Zhou, Weizhong Zhang, and Cheng Jin.
\newblock Enhancing object coherence in layout-to-image synthesis.
\newblock \emph{arXiv preprint arXiv:2311.10522}, 2023{\natexlab{b}}.

\bibitem[Wang et~al.(2024{\natexlab{a}})Wang, Zhang, and Jin]{wang2024magicface}
Yibin Wang, Weizhong Zhang, and Cheng Jin.
\newblock Magicface: Training-free universal-style human image customized synthesis.
\newblock \emph{arXiv preprint arXiv:2408.07433}, 2024{\natexlab{a}}.

\bibitem[Wang et~al.(2024{\natexlab{b}})Wang, Zhang, Zheng, and Jin]{wang2023high}
Yibin Wang, Weizhong Zhang, Jianwei Zheng, and Cheng Jin.
\newblock High-fidelity person-centric subject-to-image synthesis.
\newblock In \emph{CVPR}, 2024{\natexlab{b}}.

\bibitem[Wang et~al.(2024{\natexlab{c}})Wang, Zhang, Zheng, and Jin]{wang2024primecomposer}
Yibin Wang, Weizhong Zhang, Jianwei Zheng, and Cheng Jin.
\newblock Primecomposer: Faster progressively combined diffusion for image composition with attention steering.
\newblock \emph{arXiv preprint arXiv:2403.05053}, 2024{\natexlab{c}}.

\bibitem[Wu et~al.(2019)Wu, Zhang, Liu, Han, Liu, Ding, and Bai]{wu2019editing}
Liang Wu, Chengquan Zhang, Jiaming Liu, Junyu Han, Jingtuo Liu, Errui Ding, and Xiang Bai.
\newblock Editing text in the wild.
\newblock In \emph{ACM MM}, pages 1500--1508, 2019.

\bibitem[Xu et~al.(2021)Xu, Zhang, Wang, Price, Wang, and Shi]{xu2021rethinking}
Xingqian Xu, Zhifei Zhang, Zhaowen Wang, Brian Price, Zhonghao Wang, and Humphrey Shi.
\newblock Rethinking text segmentation: A novel dataset and a text-specific refinement approach.
\newblock In \emph{CVPR}, pages 12045--12055, 2021.

\bibitem[Yang et~al.(2020)Yang, Huang, and Lin]{yang2020swaptext}
Qiangpeng Yang, Jun Huang, and Wei Lin.
\newblock Swaptext: Image based texts transfer in scenes.
\newblock In \emph{CVPR}, pages 14700--14709, 2020.

\bibitem[Zhang et~al.(2018)Zhang, Isola, Efros, Shechtman, and Wang]{zhang2018unreasonable}
Richard Zhang, Phillip Isola, Alexei~A Efros, Eli Shechtman, and Oliver Wang.
\newblock The unreasonable effectiveness of deep features as a perceptual metric.
\newblock In \emph{CVPR}, pages 586--595, 2018.

\bibitem[Zhao and Lian(2024)]{UdiffText}
Yiming Zhao and Zhouhui Lian.
\newblock Udifftext: A unified framework for high-quality text synthesis in arbitrary images via character-aware diffusion models.
\newblock In \emph{ECCV}, 2024.

\end{thebibliography}
}
\clearpage % \section*{Appendix}
\appendix

% \subsection{Baselines}
% We compare our method against the state-of-the-art baselines including the GAN-based method: MOSTEL \cite{MOSTEL} and Diffusion-based methods: Stabel Diffusion-inpainting (v2.0) \cite{rombach2022high}, DiffSTE \cite{DiffSTE}, TextDiffuser \cite{TextDiffuser}, and UDiffText \cite{UdiffText}.
\begin{figure*}[ht]
\centering
\includegraphics[width=1\textwidth]{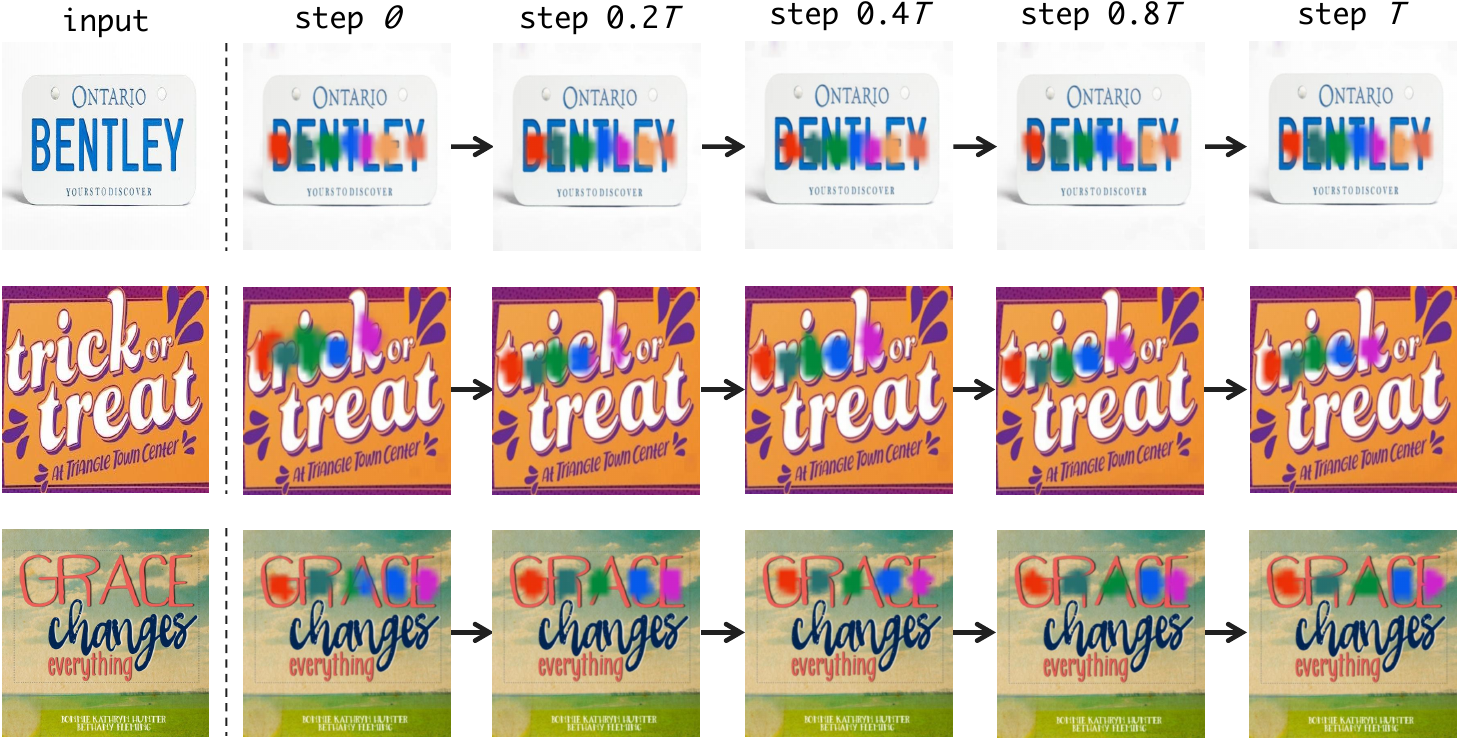}
\caption{Additional visualized attention cases of all characters across several steps during training.}
\label{fig:more_mid}
\end{figure*}

\section{Experimental Settings for Baselines} \label{appendix: exp_set}
In this study, we conduct a comprehensive comparison of our method against several state-of-the-art baselines, encompassing both GAN-based and diffusion-based approaches. Specifically, we evaluate our method against MOSTEL \cite{MOSTEL}, Stable Diffusion-inpainting (v2.0) \cite{rombach2022high}, DiffSTE \cite{DiffSTE}, TextDiffuser \cite{TextDiffuser}, AnyText \cite{tuo2023anytext} and UDiffText \cite{UdiffText}.
For MOSTEL, we utilize it to generate text within the masked region and then integrate the output back into the original image. Regarding Stable Diffusion, we employ the publicly available pre-trained model "stable-diffusion-2-inpainting" from Hugging Face, setting its prompt as ``[word to be rendered]'' for a fair comparison.
For TextDiffuser, we utilize their inpainting variant, where the desired text is rendered in a standard font (Arial) within the masked region, serving as input for their proposed segmentor. Finally, for UDiffText, AnyText, and DiffSTE, we follow the settings outlined in their respective original papers.
\begin{figure}[!t]
\centering
\includegraphics[width=1\linewidth]
{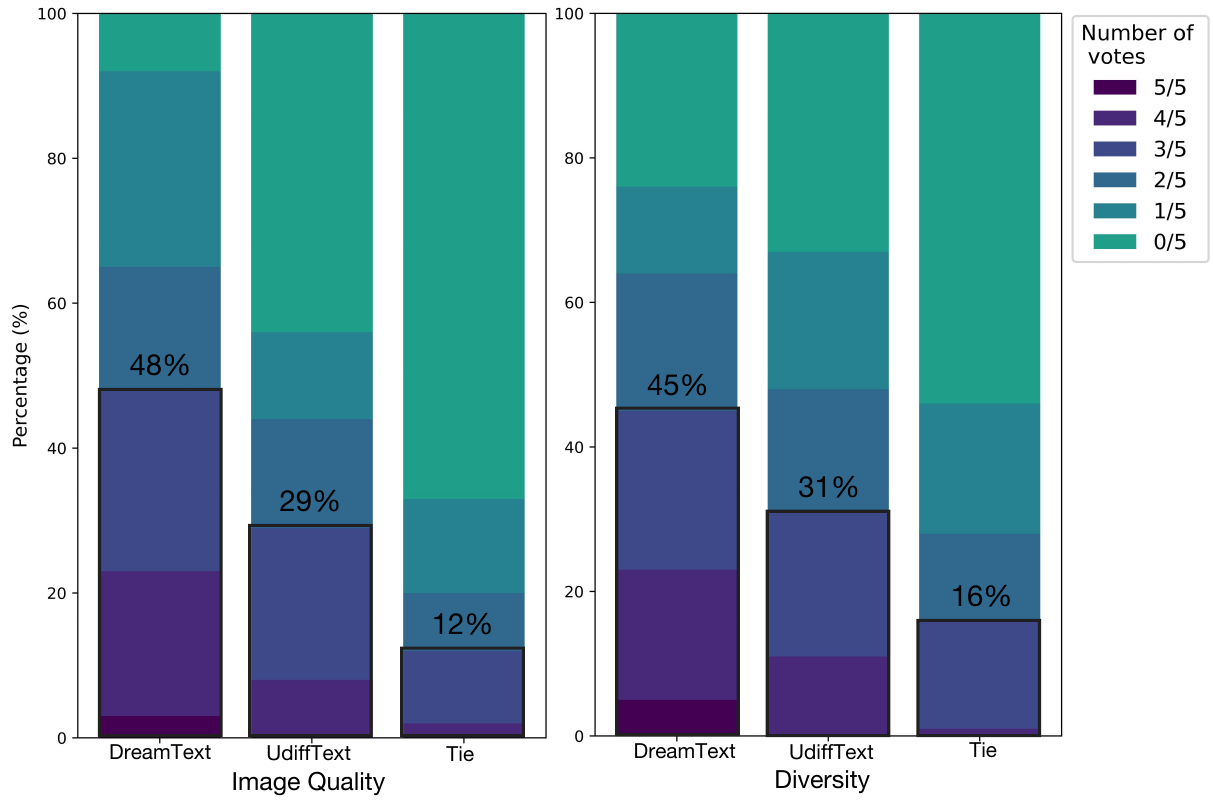}
\caption{Visualized results of human study.}

\end{figure}

\section{Ablation Study on Balanced Supervision}
We analyze the efficacy of our balanced supervision for character attention by comparing it with unsupervised and self-supervised learning approaches. We additionally incorporate the mean Intersection over Union (mIoU) metric to evaluate the alignment between their latent character masks and ground truth character segmentation masks on the LAION-OCR dataset. As illustrated in Tab. \ref{table:ablation_semi}, when conducting unsupervised learning, we observed a significant deterioration in the model's performance across all metrics. This outcome is reasonable since the model lacks prior knowledge about determining the ideal character locations at the outset. Consequently, it fails to concentrate attention effectively around character regions, leading to significant deviations between the characters' positions in generated latent character masks and their actual positions. As a result, the autonomous alternate optimization process is adversely affected.
\begin{table}[t]  
	\small
        \centering
        \setlength{\tabcolsep}{4pt}
	\caption{Ablation study results on balanced supervision.}
        
	\begin{tabular}{c|cc|cc}
		\bottomrule
		\multirow{2}{*}{\textbf{Setting}} & \multicolumn{2}{c|}{\textit{\textbf{Average SeqAcc}}} & \\
            & Recon & Editing &FID &mIoU\\
		\bottomrule
            unsupervised                  & 0.212      &0.157    & 62.36   & 0.203\\
		  supervised      &0.862      & 0.813     & 14.92 &  0.617\\
    	  balanced supervision (\textbf{Ours})         & \textbf{0.940}    & \textbf{0.887} & \textbf{12.13} & \textbf{0.722}\\
		% +$\mathcal{L}_{align}$      &0.873      & 0.712    & 26.53& \\
		% +$\mathcal{L}_{id}$           &\textbf{0.940}   &\textbf{0.847}  & \textbf{15.79} &  \\
		\bottomrule
	\end{tabular} \\
        \label{table:ablation_semi}
	% \label{table:ablation}

\end{table}

Furthermore, we explore self-supervised learning by guiding attention calibration using the cross-entropy objective between latent character masks and character segmentation masks. Although self-supervised learning demonstrates a notable improvement compared to unsupervised learning, there remains a gap when compared to our balanced supervision. This discrepancy arises from the overly strong constraint of the characters' position, limiting the model's flexibility in estimating optimal positions, which hinders its ability to adapt to varied and complex scenarios.
% impacting the model's autonomous learning to estimate the optimal character-generated position.  
% As a result, the deflected attention toward specific character regions occurs during inference, thereby constraining its overall performance.

\section{Human Study}
We conduct a human study to compare our method with UdiffText. The results are visualized in the accompanying figure. A total of 50 cases were prepared, with each case generating four images using both methods to evaluate diversity. Additionally, one image per case was randomly selected for quality assessment. We report the percentage of queries receiving positive votes, with a black box highlighting the cases where the majority consensus was achieved.

\section{Additional Visualized Attention Results}
Additional visualized attention cases of all characters across several steps during training are exhibited in Fig. \ref{fig:more_mid}.

\section{Additional Qualitative Comparative Results}
\begin{figure*}[ht]
\centering
\includegraphics[width=0.7\textwidth]{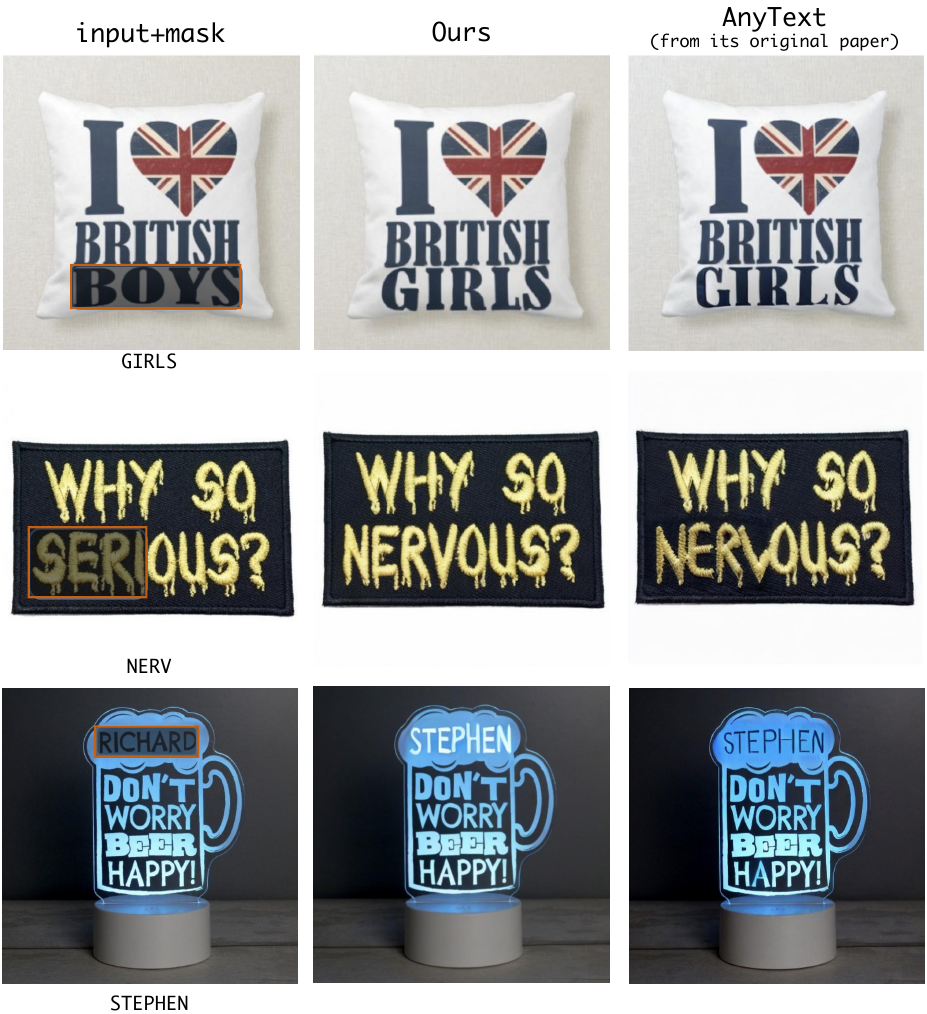}
\caption{Qualitative comparative results against AnyText \cite{tuo2023anytext}.}
\label{fig:compare_anytext}
\end{figure*}

\begin{figure*}[ht]
\centering
\includegraphics[width=\textwidth]{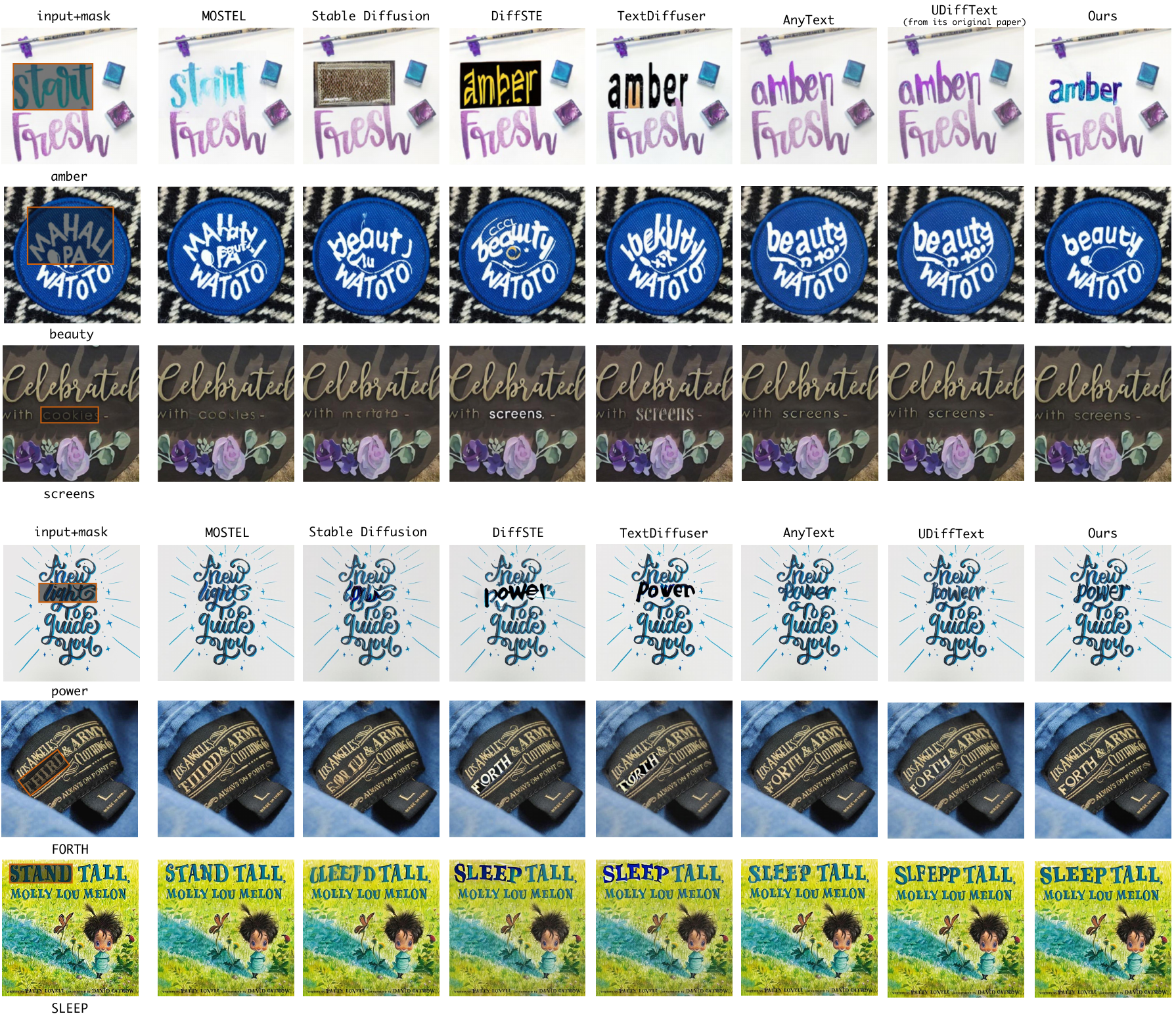}
\caption{Additional qualitative comparative results against state-of-the-art methods.}
\label{fig:more_compare}
\end{figure*}

Additional qualitative comparisons against the baselines are exhibited in Fig. \ref{fig:compare_anytext} and \ref{fig:more_compare}.

\section{Additional Visual Results}
\begin{figure*}[ht]
\centering
\includegraphics[width=0.9\textwidth]{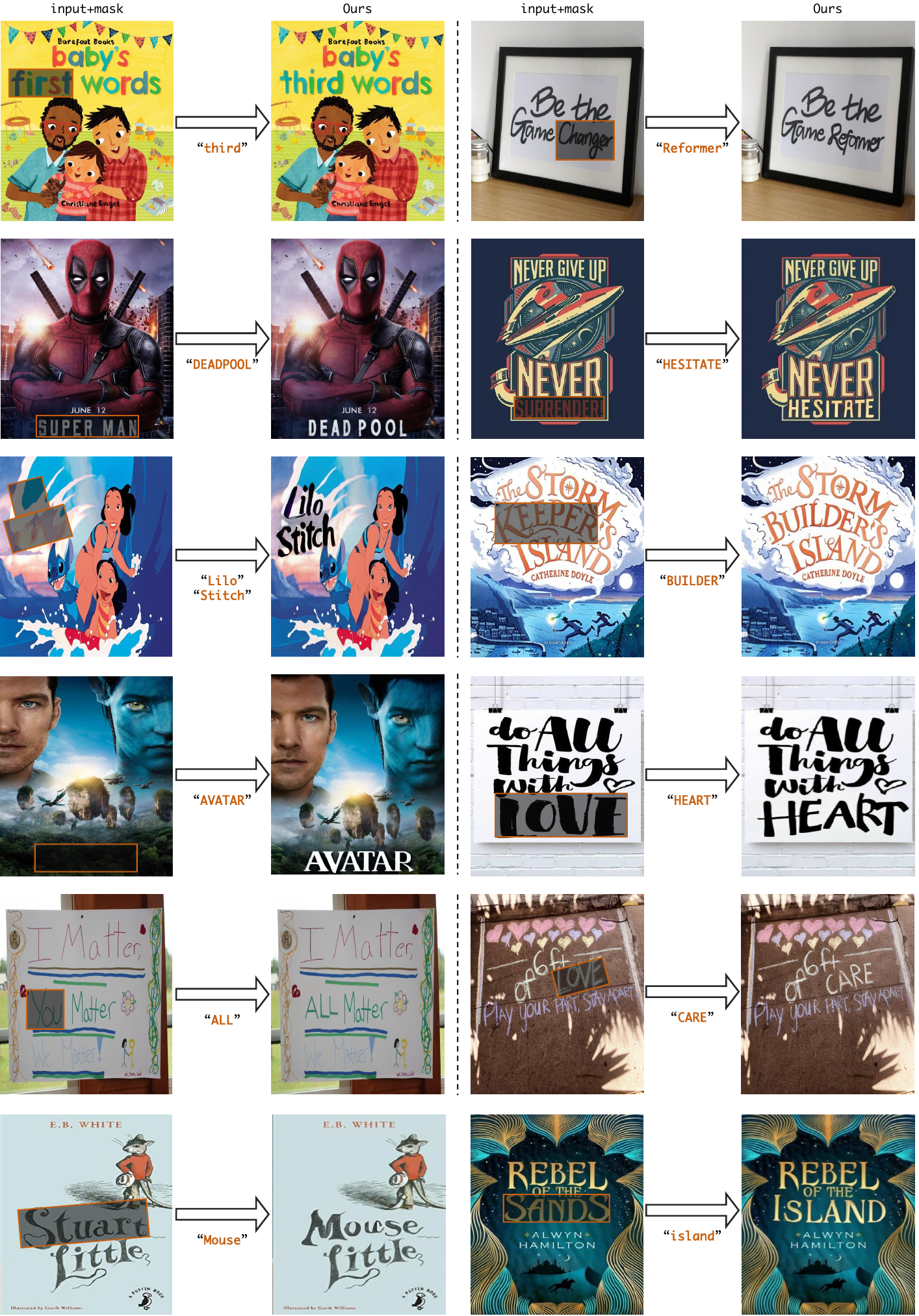}
\caption{Additional visual results generated by our \ourmethod.}
\label{fig:more_display1}
\end{figure*}

\begin{figure*}[ht]
\centering
\includegraphics[width=0.9\textwidth]{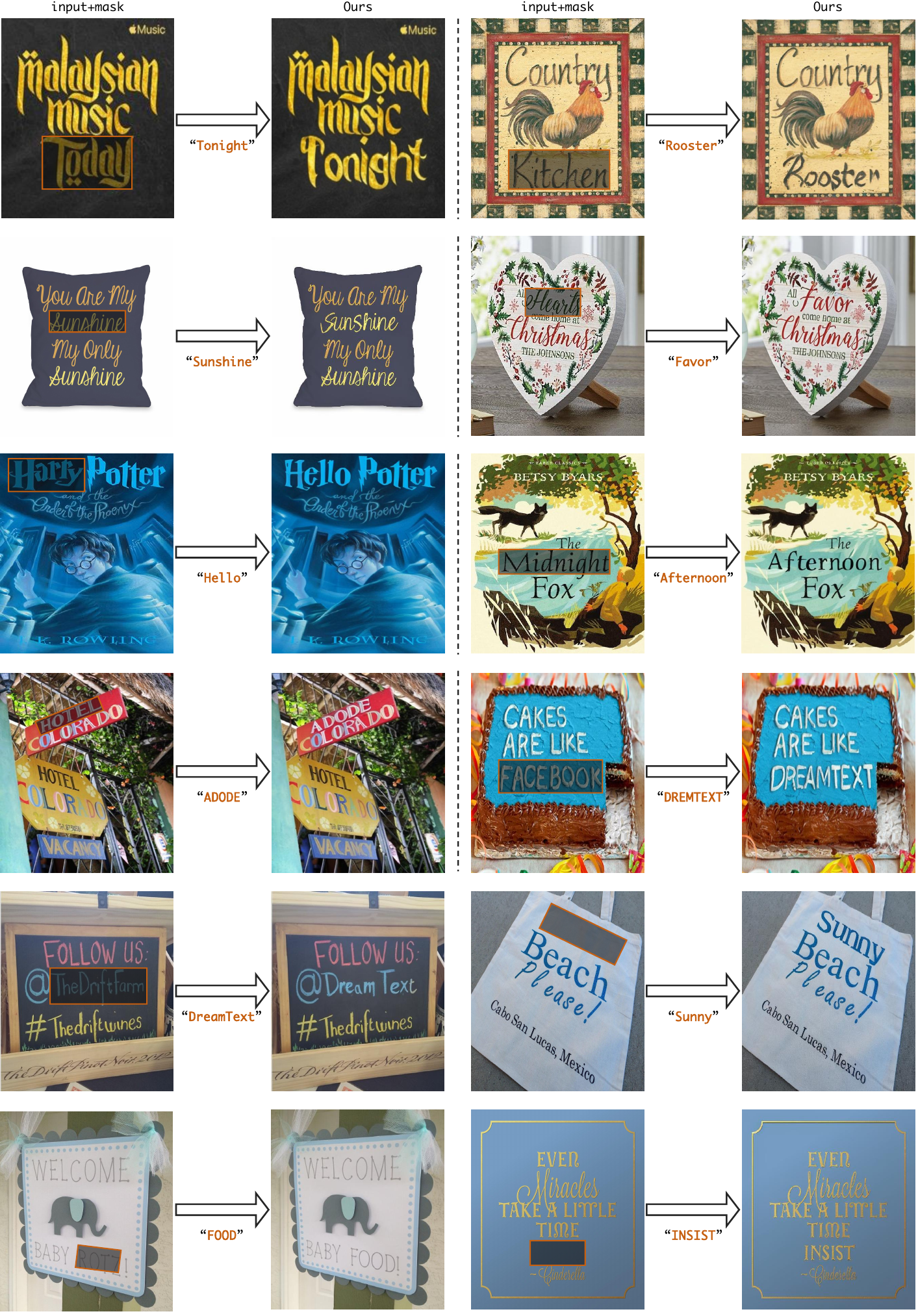}
\caption{Additional visual results generated by our \ourmethod.}
\label{fig:more_display2}
\end{figure*}

Additional visual results generated by our \ourmethod are exhibited in Fig. \ref{fig:more_display1} and Fig. \ref{fig:more_display2}.


\section{Limitations} \label{limitation}
While our method demonstrates promising capabilities in synthesizing scene text, it is limited by its inability to simultaneously modify multiple regions within an image, which restricts its applicability. Future research will explore techniques to address this limitation, aiming to develop more efficient and versatile text synthesis methods capable of simultaneously generating multiple texts within an image. Besides, its application prompts considerations regarding privacy. The generation of realistic text, including personal signatures or identifiable information, may pose risks if misused, potentially compromising individuals' privacy and security. These concerns underscore the importance of implementing robust safeguards and ethical guidelines to address potential privacy risks and ensure the responsible use of this technology.

\section{Societal Impact} \label{societal_impact}
The advancement of scene text synthesis technology in our work holds significant societal implications. Firstly, it contributes to cultural preservation by enabling the generation of text in diverse styles and languages, aiding in the digitization and conservation of historical scripts and manuscripts. Additionally, our method has applications in art, design, and advertising, empowering creators to produce visually captivating compositions and typography designs. However, ethical considerations surrounding the potential misuse of synthesized text for fraudulent purposes must be carefully addressed through the development of robust safeguards and guidelines. Therefore, while our work offers promising possibilities for text synthesis, it necessitates thoughtful consideration of its societal impacts and ethical implications.

\section{Ethical Statement}
In this work, we affirm our commitment to ethical research practices and responsible innovation. To the best of our knowledge, this study does not involve any data, methodologies, or applications that raise ethical concerns. All experiments and analyses were conducted in compliance with established ethical guidelines, ensuring the integrity and transparency of our research process.

% \clearpage \input{8_appendix}

\end{document}